\documentclass[journal]{IEEEtran}
\usepackage[utf8]{inputenc}
\usepackage[T1]{fontenc}
\usepackage{xcolor,soul,framed} 
\usepackage[pdftex]{graphicx}
\graphicspath{{../pdf/}{../jpeg/}}
\DeclareGraphicsExtensions{.pdf,.jpeg,.png}
\usepackage{amsmath}
\usepackage{eqparbox}
\usepackage{url}
\usepackage[noadjust]{cite}
\usepackage{multicol}
\usepackage{amssymb}
\usepackage{breqn}
\usepackage[caption=false, font=footnotesize]{subfig}
\usepackage{array}
\usepackage{float}
\usepackage{makecell}
\usepackage[ruled,linesnumbered]{algorithm2e}
\usepackage{algorithmic}
\usepackage{mathrsfs}
\usepackage{mathtools, color}

\DeclareMathOperator*{\argmin}{arg\,min}
\DeclareUnicodeCharacter{2212}{-}
\DeclareUnicodeCharacter{2217}{-}
\DeclareUnicodeCharacter{2248}{-}

\begin{document}
	
	%
	\title{3D Sensing of a Moving Object with a Nodding 2D LIDAR and Reconfigurable Mirrors}
	

	\author{Anindya~Harchowdhury,
	Lindsay~Kleeman,
	and~Leena Vachhani, 
		
	}
	
	\maketitle
	
	
	\begin{abstract}
		Perception in 3D has become standard practice for a large part of robotics applications. High quality 3D perception is costly. Our previous work on a nodding 2D Lidar provides high quality 3D depth information with low cost, but the sparse data generated by this sensor poses challenges in understanding the characteristics of moving objects within an uncertain environment. This paper proposes a novel design of the nodding Lidar but provides dynamic reconfigurability in terms of limiting the field of view of the sensor using a set of optical mirrors. It not only provides denser scans, it achieves three times higher scan update rate. Additionally, we propose a novel calibration mechanism for this sensor and prove its effectiveness for dynamic object detection and tracking.
	\end{abstract}
	
	\section{Introduction}
\begin{figure}[t]
		\begin{center}
			\includegraphics[height=5.cm, width=6.2cm]{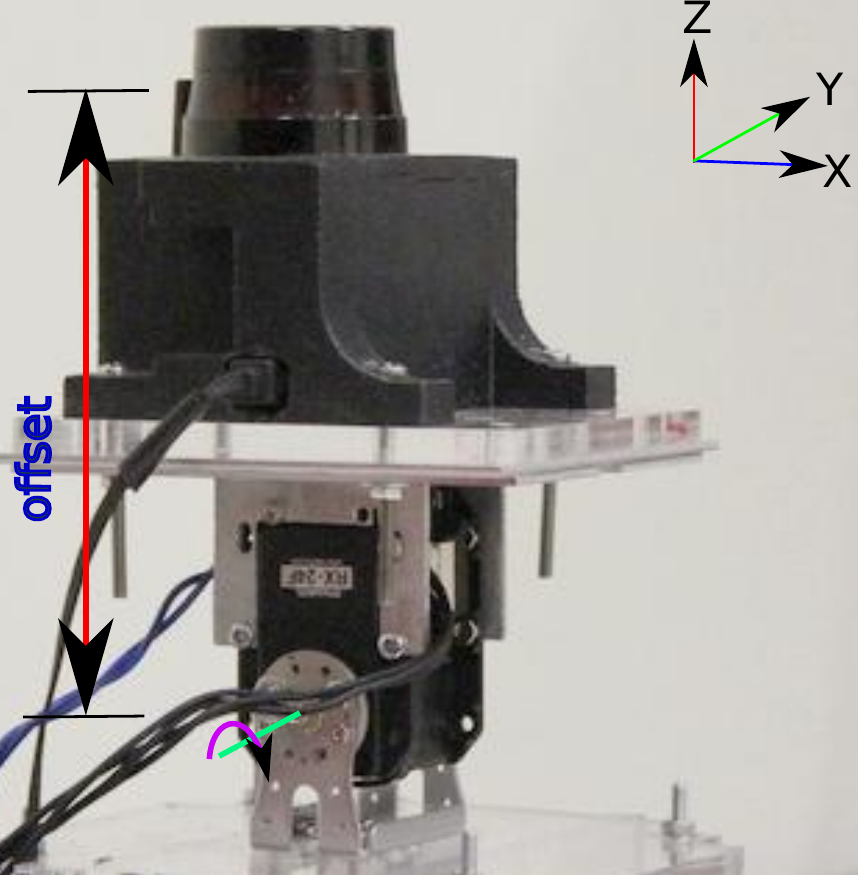}
			\caption{Transverse view of Nodding 2D Lidar}
			\label{fig:01}
		\end{center}
	\end{figure}

	\IEEEPARstart{L}{idar} is a reliable sensor for autonomous robotic platform research and development over the last two decades. Its accurate range measurement keeps it competitive with other sensory systems for navigation such as cameras and RGB-D sensors. The optoelectronics and precise motor control of commercial Lidars, make them costly. Sate-of-the-art 3D Lidars are significantly more expensive than 2D Lidars and this motivated our nodding 2D Lidar in \cite{ral18}, shown in Figure~\ref{fig:01}.
	
\begin{figure*}[t]
		\centering
		\subfloat[]{
			\includegraphics[height = 1.75in, width=0.32 \linewidth]{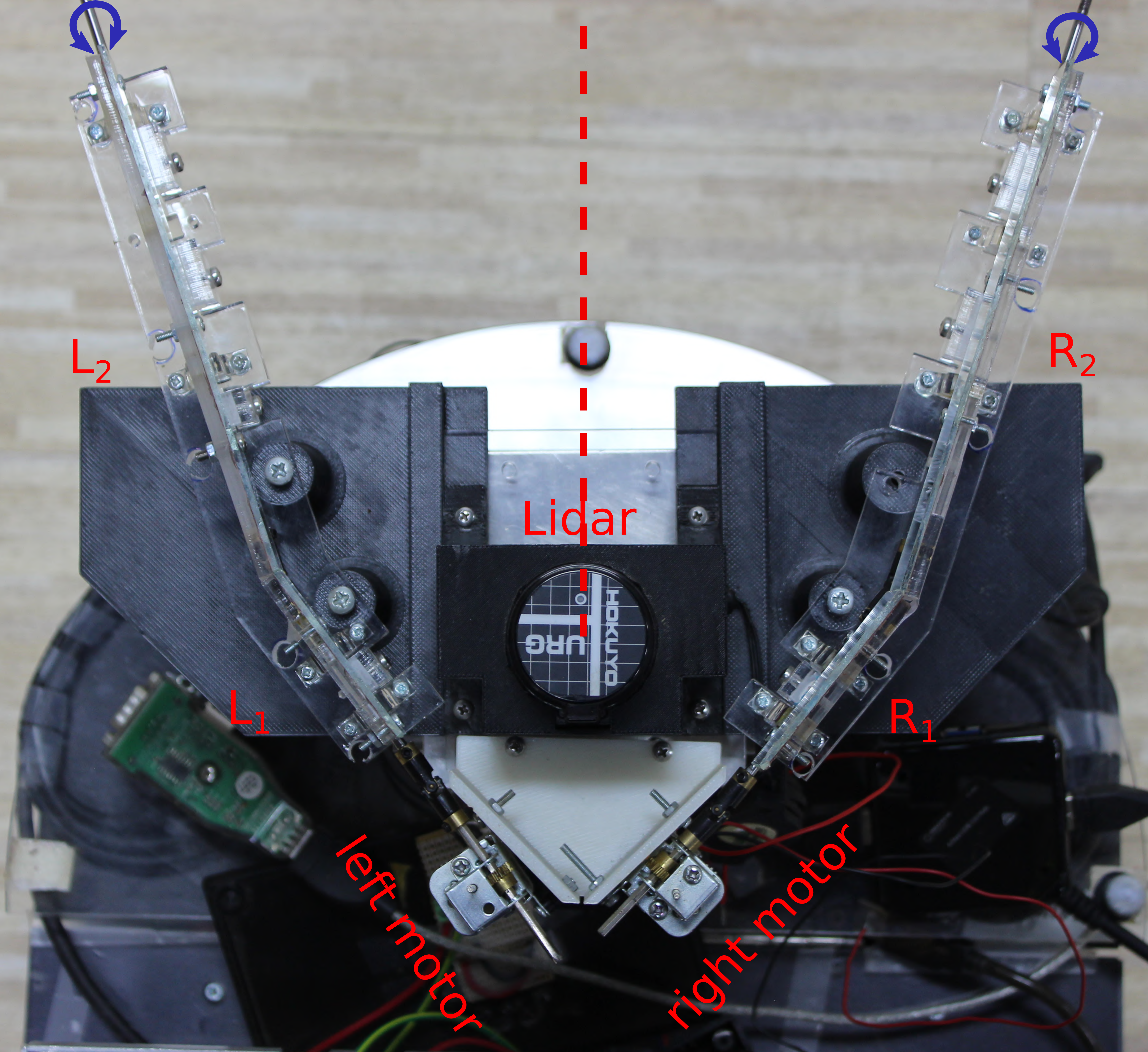}}
		\label{1a}\hfill
		\subfloat[]{
			\includegraphics[height = 1.77 in, width=0.66 \linewidth]{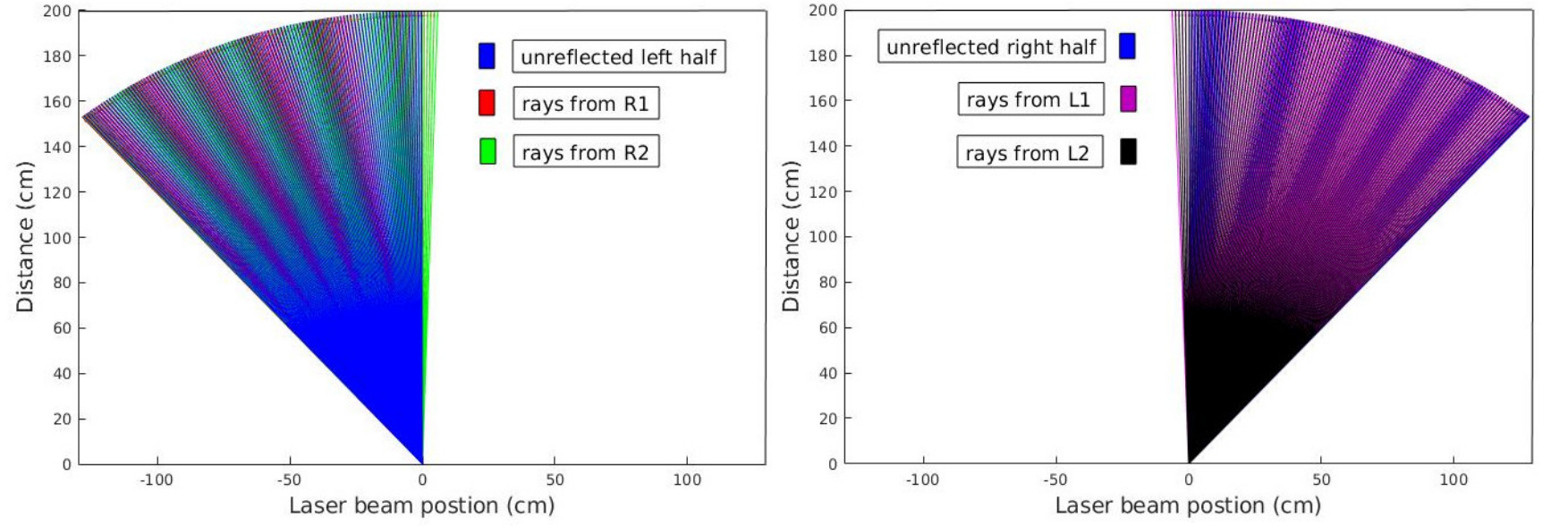}}
		\caption{(a) Reconfigurable nodding Lidar; left and right motors rotate the corresponding set of mirrors to provide reconfigurability of the scan window; the Lidar is shown in the middle; (b) for clarity in viewing, beam scan pattern due to reflected and un-reflected laser beams are separately shown in two figures.}
		\label{fig:02}
	\end{figure*}

	State-of-the-art Lidars come with a large angular Field of View (FoV). While tracking a dynamic object, it is more important to increase the frequency of scans, rather than  scanning a wide angular region consisting of many obstacles. 
	Use of optical reflectors placed in the FoV is proposed to reduce the sensor's scan window. These reflectors can be placed in the FoV of the sensor to reduce its scan window and exploit the possibility of faster scan update rate using reflected laser beams within the narrower modified FoV. The  judicious choice of mirror placement can achieve higher angular resolution by segmenting the original FoV into multiple equal parts. In this work, the original FoV of the sensor is divided into three equal parts and the incident beams are redirected at either side, using a set of mirrors. A photo of the sensor comprising of a nodding Lidar with mirrors is shown in Figure~\ref{fig:02}. This mechanism will concentrate all the laser beams, within the 1/3rd angular section in the middle, in front of the sensor as shown in Figure~\ref{fig:02}. The motorized control of the mirrors will allow any user to lift up or reset the mirror position on the fly. We consider HOKUYO URG-04LXUG01 laser range finder in this work, although our approach is not limited to any specific range sensor. Based on this idea, the parameters relevant to mirror placement are derived in Section III. 
	
	
	The rest of the paper is organized as follows. A brief study of the existing research in connection with this paper is discussed in Section II. Mathematical derivations towards obtaining the optimal mirror(s) placement and their dimensions are provided in Section III. A novel calibration mechanism necessary for optimal performance of the sensor is proposed in Section IV. Dynamic obstacle handling and its motion estimation algorithm is detailed in Section V. A number of experiments have been conducted and the analysis of the corresponding results are presented in Section VI. Finally, concluding remarks are included in Section VII, together with the challenges and contributions discussed in this paper. 
	\section{Related Works}	
	
	Extending the capability of a 2-D Lidar to scan in 3-D has been investigated in various ways. 
	A spring mounted Lidar, {\em zebedee} introduced in \cite{icra1, tro1}, shows the usefulness of nodding 2D Lidar for simultaneous localization and mapping (SLAM) problem. The capability of a spinning Lidar to produce reliable 3-D scan was shown in \cite{key1}, although the scans registered are found to be sparse in nature. In another interesting 3-D depth sensing mechanism \cite{icra6}, an electromagnetically vibrating gold coated glass mirror has been mounted on a rotating  motor to work like a 3-D depth sensor. A rigorous study on actuating Lidars in \cite{ROB}, shows the performance of a rotating 2D Lidar about all the elementary reference axes. The results of this work demonstrated the impact of each type of actuation on the scanning performance. A very recent work presented in \cite{hin01}, shows a low-cost rotating Lidar based 3-D perception system, in that a  calibration mechanism for this kind of sensor is discussed, and it has been shown to register point cloud with reasonable accuracy while mounted on a mobile robot. In all these works, the sensor comes out with its factory settings and  does not provide any reconfigurability of the FoV, while using all its resources. Particularly for a 2-D Lidar, while used for navigation in unknown environment with dynamic objects, faster measurement update along with denser depth data are expected, and that inspires us to investigate on a reconfigurable nodding Lidar.

	There are variety of range based object tracking algorithms, that use depth characteristic of dynamic object(s) for motion estimation.
	Representing an environment with static and dynamic objects, using voxels have been recently found in \cite{spr15}. One of the ground braking work on 3-D mapping \cite{hor}, can be useful in memory efficient representation of dynamic obstacle(s). Lidar based dynamic object tracking performance has been improved both in speed and accuracy using octree based indexing in \cite{iros16_2}. Model independent six-DOF motion estimation of dynamic objects using range image has been discussed in \cite{icra10_3}. Segmenting out large background static objects is an important preliminary task for motion estimation. Two recent works  \cite{iros08_2, iros12_2} show robust point cloud segmentation, and a fast plane fitting from noisy range image in 3-D, that can be used for static object segmentation. We take inspiration from the above works and propose a simple solution to segment a dynamic object and estimate its motion parameters. 

	
	
	
	
	%

	\section{Mathematical Model of Sensor set-up}
	
	Originally, the Lidar has $240^{\circ}=\frac{4\pi}{3}$rad FoV in its scan plane, and $I=682$ valid range measurements, so that upon using mirrors, the modified angular window size becomes $\frac{4\pi}{9}$rad. Laser beams transmitted from the $\frac{4\pi}{9}$rad window in the middle are unreflected and directly impinge on objects and the rest of the beams from other two windows are reflected and redirected within the $\frac{4\pi}{9}$rad scan region in the front of the sensor. Due to geometric constraints and infeasible size requirements of the mirrors, one mirror at each side is not enough to satisfy the objective. Two mirrors at each side can do the job, but cannot illuminate the whole $\frac{4\pi}{9}$rad angular window in the front. The mirrors in each side illuminates half of the angular window at the other side about the perpendicular drawn from sensor origin. Four mirrors, namely $L_1$ and $L_2$ at the left, and $R_1$ and $R_2$ at the right are shown in Figure~\ref{fig:02}.
	The sensor scans the environment in anti-clockwise manner. Therefore, based on our mirror placement, $L_2$ is followed by $L_1$ in terms of generating reflected signals. The opposite happens at the other side of the sensor with the help of mirrors $R_1$ and $R_2$. As 1024 laser beams are swept in $2\pi$ rad, angular resolution of this Lidar is $\frac{\pi}{512}$rad. The objective is to fill the space between two consecutive unreflected beams with the specular reflected beams while hitting an obstacle. This way, two sets of reflected beams along with the unreflected ones would be accommodated adjacent to each other while maintaining $\frac{\pi}{1536}$rad inter-beam angular separation. The resulting scenario is graphically shown in Figure~\ref{fig:02}.

\begin{figure}[t]
		\begin{center}
			\includegraphics[height=10.9cm, width=7.7cm]{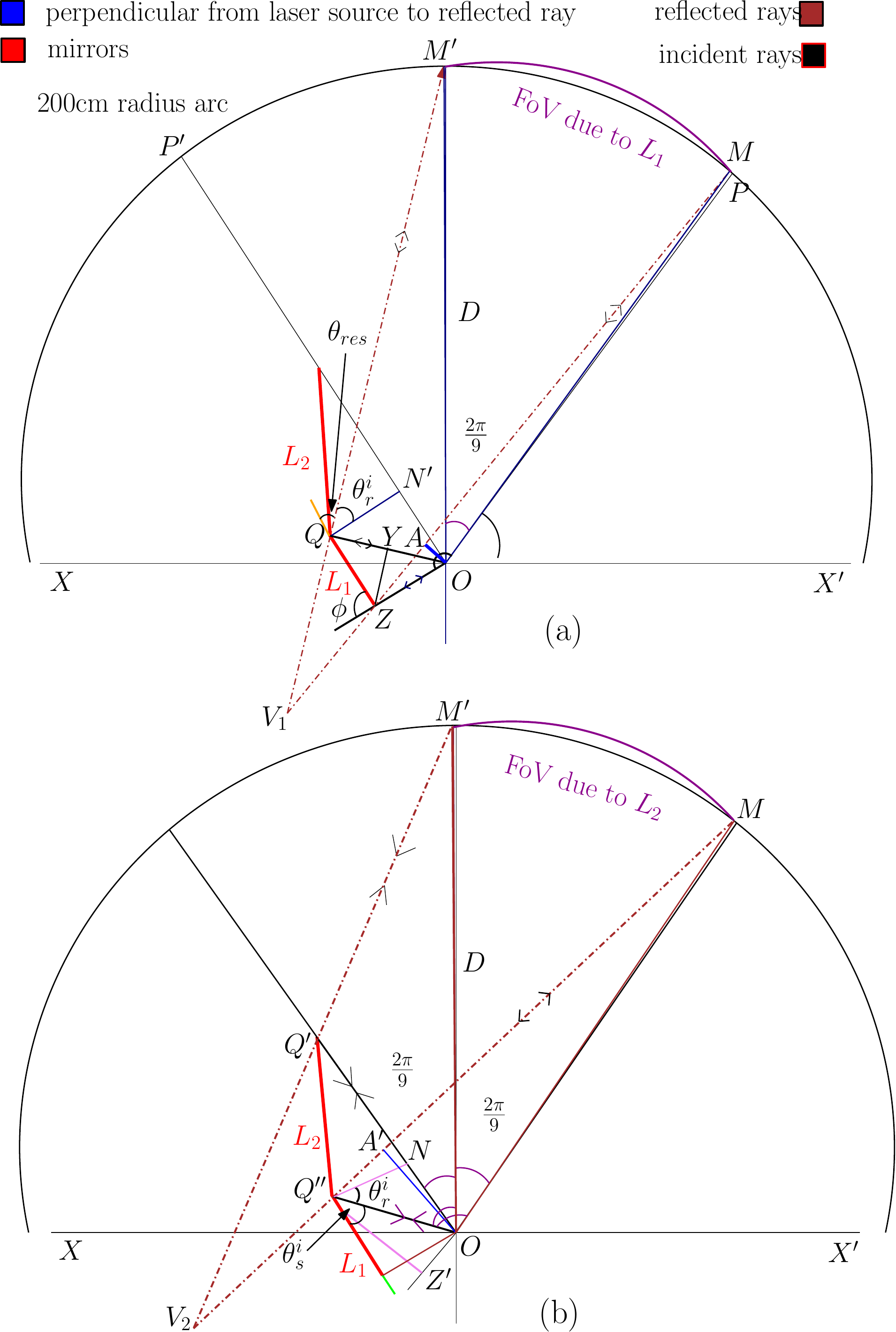}
			\caption{Un-scaled version of mirror configuration; for (a) $L_1$ and (b) $L_2$ mirror}
			\label{fig:04}
		\end{center}
	\end{figure}

	Here, we provide the mathematical formulation to compute the required design parameters for the optimal placement of the mirrors, that will generate the desired scan pattern. Due to large range variation compared to the size of mirrors, it is difficult to draw a scaled diagram. To bring clarity for the geometric description of mirror $L_1$, an un-scaled version is presented in Figure~\ref{fig:04}a. Here, center of the sensor is denoted by $O$.
	Let the measured locations be situated on an arc with radius $D$. From $\triangle ZOM$, $OM = D$. The last beam incident on mirror $L_1$ is $OZ$ and the corresponding reflected beam is denoted by $ZM$. The virtual image is formed at $V_1$. The length of $L_1$ is represented by $QZ$. The angle at which the mirror is subtended on $OZ$ is denoted by $\phi$. The angle of reflection corresponding to $i^{th}$ transmitted laser beam incident on the mirror is denoted by $\gamma_i$, which is for example $\angle N'ZM$ with respect to the incident beam $OZ$ on $L_1$. Horizontal reference axis is denoted by $XX'$. The bearing of the incident laser beams are known. Considering the reflected beam $ZM$, the actual distance between the sensor and $M$ is $OM$. $OP$ and $OP'$ are the two extreme unreflected beams emanating from the sensor. As discussed above, to place two reflected beams between two consecutive unreflected beams while hitting on an object,our aim is, $\angle X'OM$ to be at $\frac{\pi}{1536}$rad offset to the right from the unreflected first beam $OP$, while viewed from $O$. Therefore, $\angle X'OM$ is known to us. $X'OZ$ is the bearing angle for the $I$-th beam. 
	
	In $\triangle ZOM$, $\angle OMZ = \arcsin( \frac{OM}{OA})$. Therefore, $\angle OZM = \pi − (\angle MOZ + \angle OMZ)$. Then, $OZ = OA \sin(\angle OZM )$.  $\angle OZM$ is the sum of angle of incidence and angle of reflection for the incident beam $OZ$. Then, for $i^{th}$ incident beam $\gamma_i = \angle OZM/2$. The complementary angle corresponding to $\gamma _r^i$ is denoted by $\gamma _c^i$ and expressed as, $\gamma _c^i = ( \frac{\pi}{2} − \gamma_i)$. To calculate $\phi$, we use the $I$-th beam. Then, $\phi = \pi − (\gamma_c^{I} + 2\gamma_r^{I})$. Also, $\angle OQZ = \pi − (\gamma_c^{I} + 2\gamma_r^{I} + \angle QOZ)$. Now, the first beam to hit on $L_1$ is OQ and due to the knowledge of bearings of the transmitted laser pulses, we can calculate $\angle QOZ$ as $(\angle X'OZ − \angle X'OQ)$. At this point, let us draw a perpendicular $YZ$ on $OQ$ from $Z$. From $\triangle OYZ$, $YZ = OZ \sin(\angle YOZ)$. Then, from $\triangle OYZ$, $OY = \frac{YZ}{\tan(\angle YOZ)}$ and from $\triangle QYZ, YQ = \frac{YZ}{\tan(\angle OQZ)}$. Now the length of $L_1$, $QZ = \frac{YZ}{\sin(\angle YQZ)}$. Then, we refer to Figure~\ref{fig:04}b, for deriving the parameters relevant to $L_2$. The last beam to hit mirror $L_2$ is $OQ''$ which is the immediate one to the beam $OQ$ hitting on $L_1$. The first beam hitting on $L_2$ is $OQ'$. 
	Similar approach can be adopted to calculate the length and orientation of $L_2$. As the sensor provides range and bearing information, given a reflected range measurement corresponding to a bearing angle, we can calculate the actual angular position of the object related to the measurement. From the above formulation, actual angular position $\theta_{a}$ of an object hit by a mirror reflected beam with reference to Figure~\ref{fig:04} (e.g. $\angle X'OM$) can be obtained as,
	\begin{equation}\label{eq:01}
	\theta_{a}^i = \theta_{b}^i + 2\theta_{r}^i + \arccos\bigg(\frac{AM}{D}\bigg) - \pi, \text{when $ZM$} > \text{$ZA$}
	\end{equation}
	where, $ZA = OZ ∗\cos(2\theta_r^i)$ and $\theta_b^i$ is the bearing angle of the $i$-th incident beam as $\angle X'OZ$ for $OZ$ in Figure~\ref{fig:04} and $\theta_r^i$ is the corresponding angle of reflection due to the mirror.
	
	Now, we can obtain the dependence of $D$ on $\theta_{a}$ using \eqref{eq:01}. As the distance of the measured location varies from $D$ to $D^*$, the difference in the actual bearing angle while measured at different distance is give by,
	\begin{equation}\label{eq:02}
	\theta - \theta^* = \arccos \bigg(\frac{AM}{D}\bigg) - \arccos \bigg(\frac{AM^*}{D^*}\bigg)
	\end{equation}
	where, $AM^*$ varies based on $D^*$ and can be obtained from the measured range values. Further, it is evident that the separation of angular position between two consecutive reflected beams vary over the distance and it is given as,
	\begin{dmath}\label{eq:03}
		\Delta \theta - \Delta \theta^* = \bigg(\arccos\bigg(\frac{AM_1}{D}\bigg) - \arccos\bigg(\frac{AM_1^*}{D^*}\bigg)\bigg) - \\
		\bigg(\arccos\bigg(\frac{AM_2}{D}\bigg) - \arccos\bigg(\frac{AM_2^*}{D^*}\bigg)\bigg)
	\end{dmath}
	where, $\Delta\theta$ is the angular separation between two consecutive reflected beams. $AM_1$ and $AM_2$ correspond to $AM$ for two consecutive laser beams.
	
	\subsection{Performance measure}
\begin{figure}[t]
		\centering
		\subfloat[]{
			\includegraphics[height = 1.6in, width=0.48 \linewidth]{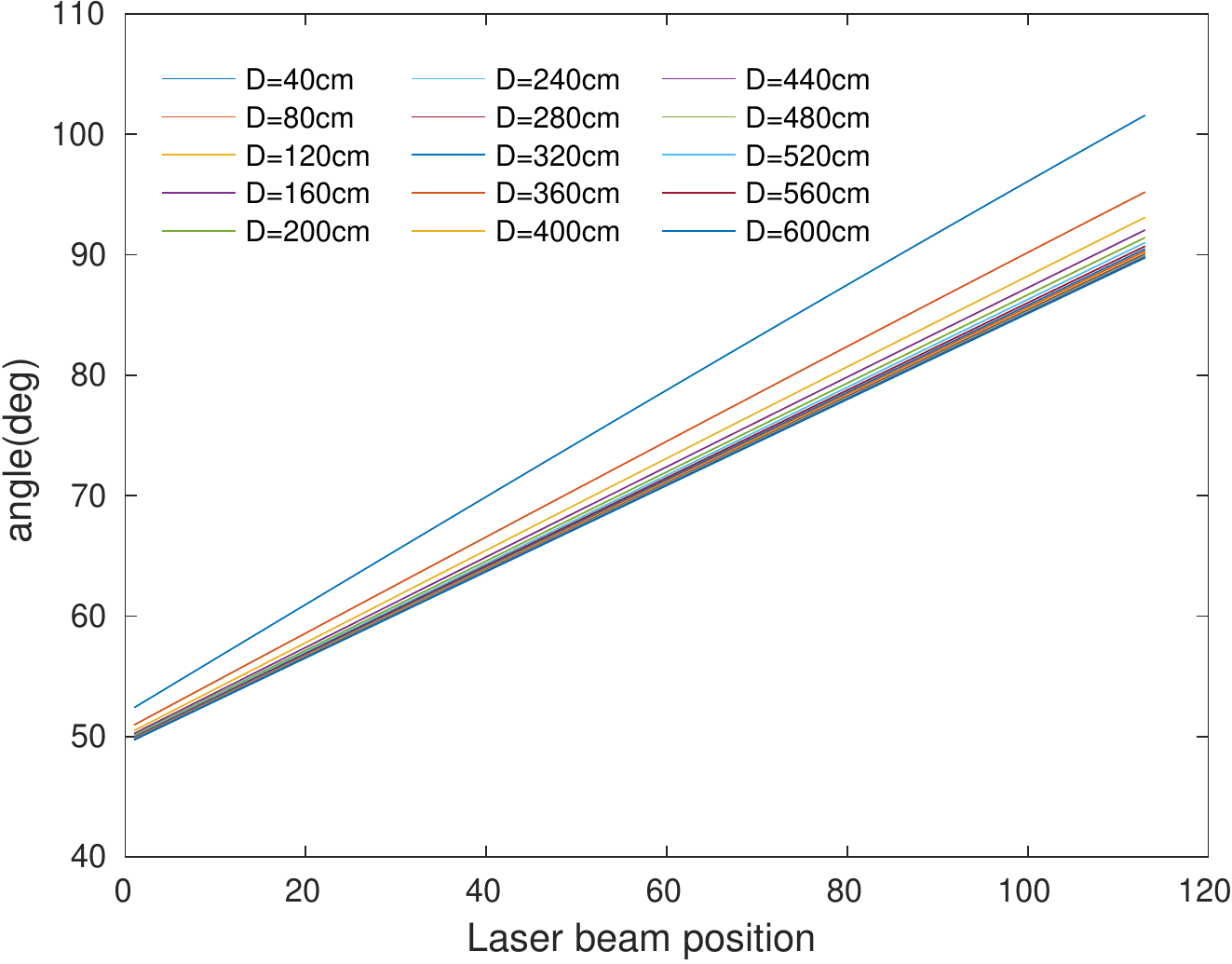}}
		\label{1c}\hfill
		\subfloat[]{
			\includegraphics[height = 1.6in, width=0.48 \linewidth]{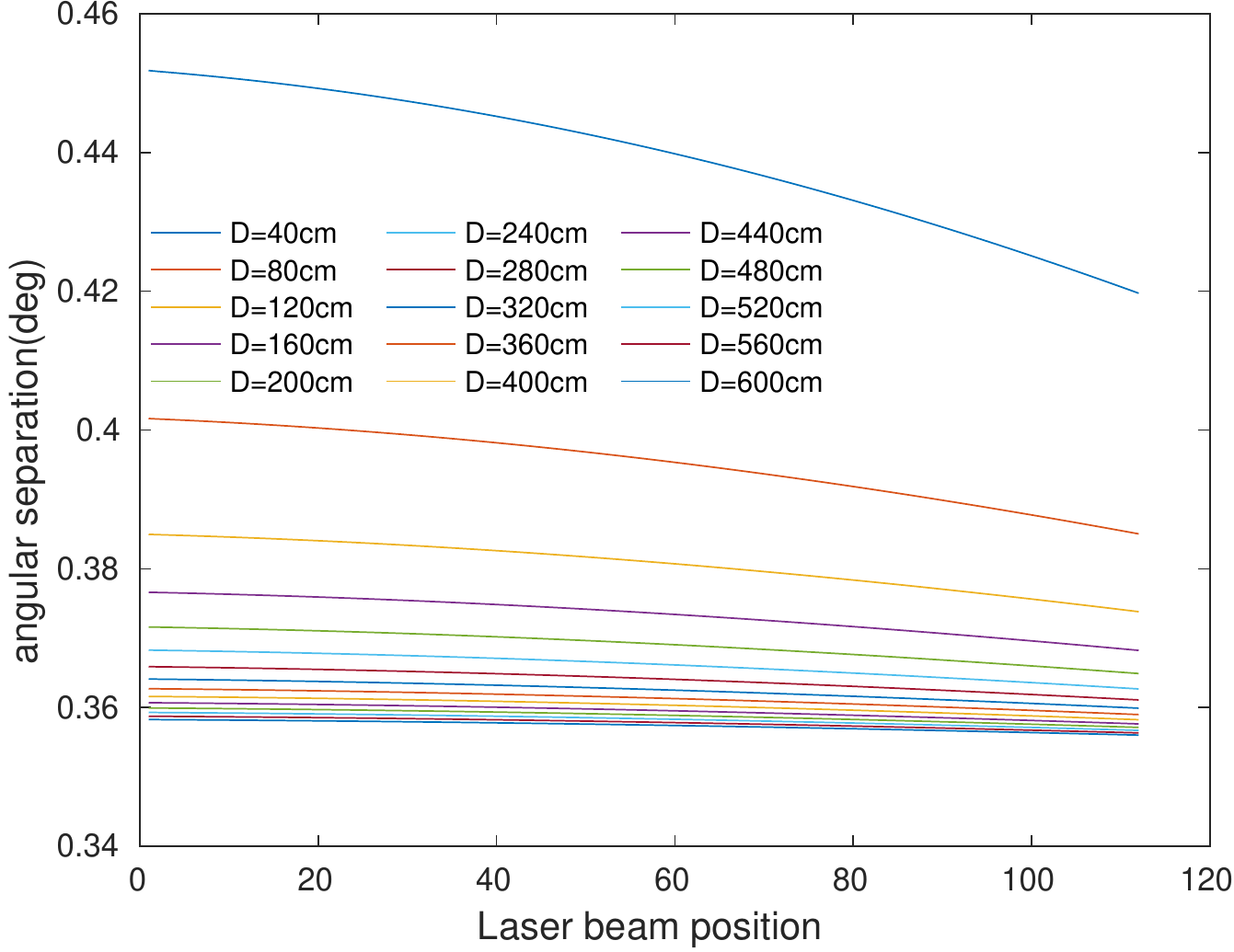}}
		\caption{(a) angular position, (b) angular separation between successive laser
			beams varied over different distances for $L_2$}
		\label{fig:06}
	\end{figure}

\begin{figure}[t]
		\centering
		\subfloat[]{%
			\includegraphics[height = 1.6in, width=0.48 \linewidth]{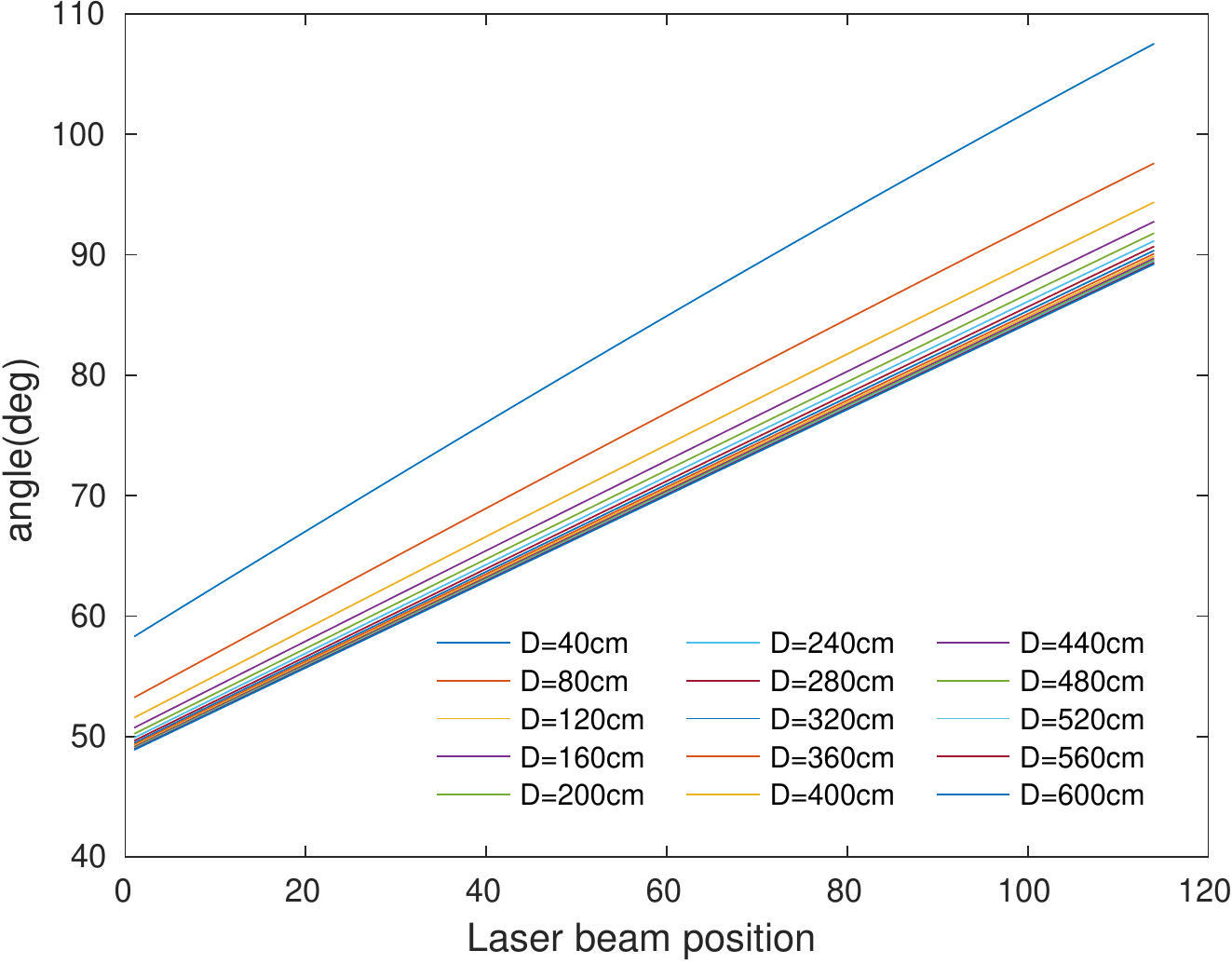}}
		\label{1f}\hfill
		\subfloat[]{
			\includegraphics[height = 1.6in, width=0.48 \linewidth]{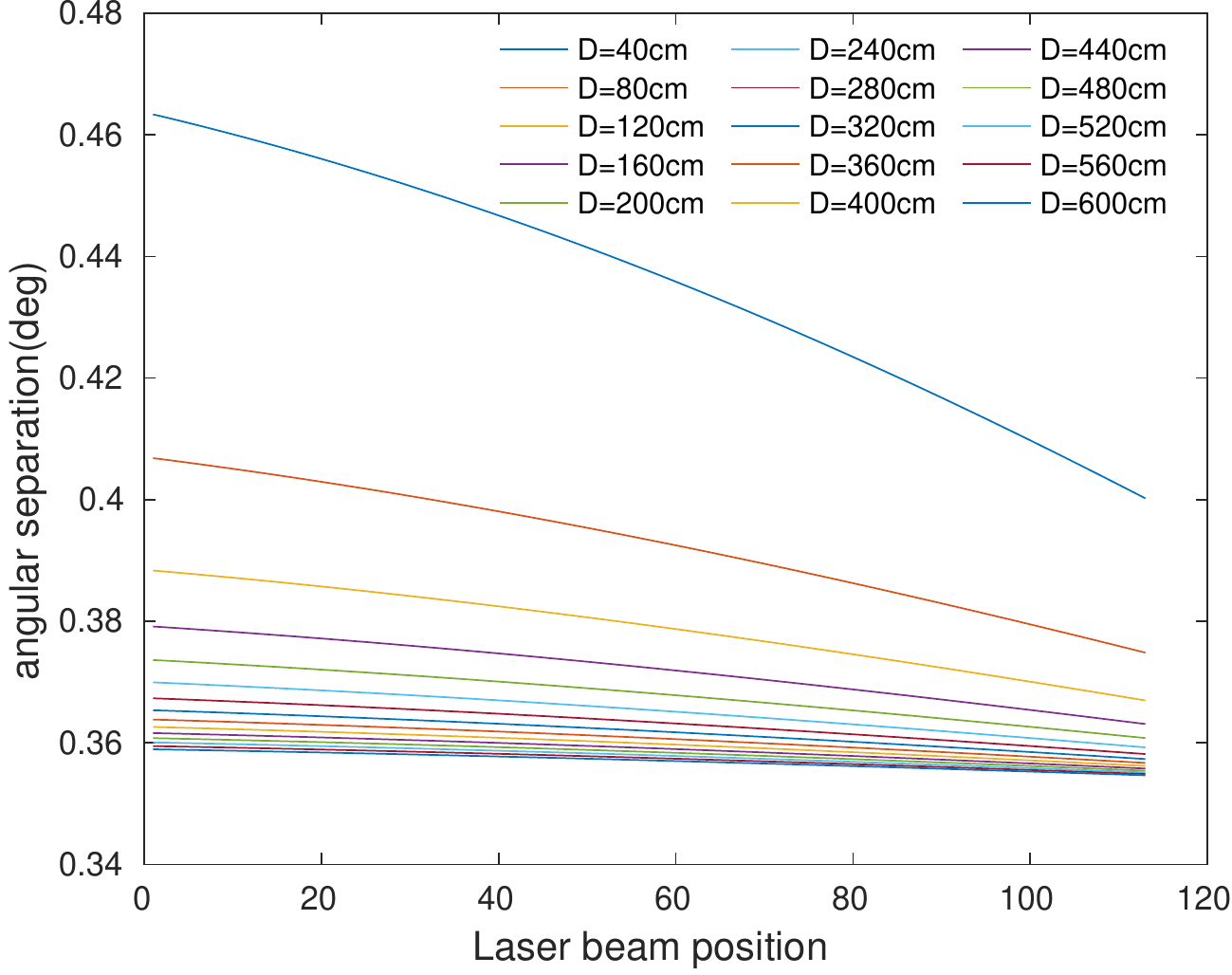}}
		\caption{In (a) angular position, (b) angular separation between successive laser
			beams varied over different distances for $L_2$}
		\label{fig:07}
	\end{figure}

	To be able to accommodate the reflected beams in the desired manner, it is necessary to set a value of $D$. We have considered $D = 200$cm, while calculating the dimension and orientation of the mirrors. Based on the formulation provided earlier in this section, the length of $L_1 = 5.4692$cm, and $\phi = 80.0397\text{deg} (1.39695\text{rad})$. For better readability, we express the angles in degrees in these results. From Figure~\ref{fig:06}, the length of $L_2 = 14.6019$cm and $\phi = 60.8307\text{deg }(1.06169 \text{rad})$. Figure~\ref{fig:06}a, and  Figure~\ref{fig:07}a present the actual bearing of the beams reflected from $L_1$ and $L_2$, and incident on an arc with different radius ranging from $D = 40$cm to $D = 600$cm. We observe that the bearing of the beams for any given arc radius varies approximately linearly and the difference between the bearing angle for the same laser beam decays drastically for increasing arc radius and we model it mathematically in \eqref{eq:02} and \eqref{eq:03}. It further shows that when $D$ is small, the bearings of the reflected beams remain close to the desired bearing angle for a given laser beam. It is evident from Figure~\ref{fig:06}b and Figure~\ref{fig:07}b that, as $D$ increases to a large number (e.g. $D=600$cm), the bearing of the reflected beams deviate largely from the desired ones. From \eqref{eq:02} and \eqref{eq:03} we find that angular separation between two neighboring beams impinging on an object depends on the distance between the sensor and the object due to the reflection caused by mirror. We demonstrate this phenomenon in Figure~\ref{fig:06}b and Figure~\ref{fig:07}b, for different $D$ from the sensor. As the distance increases the angular separation between consecutive beams gradually becomes constant.  With this characterization, the calibration procedure of the sensor is proposed next.
	
	\section{Sensor Calibration}
\begin{figure}[t]
		\begin{center}
			\includegraphics[height=4.9cm, width=8.4cm]{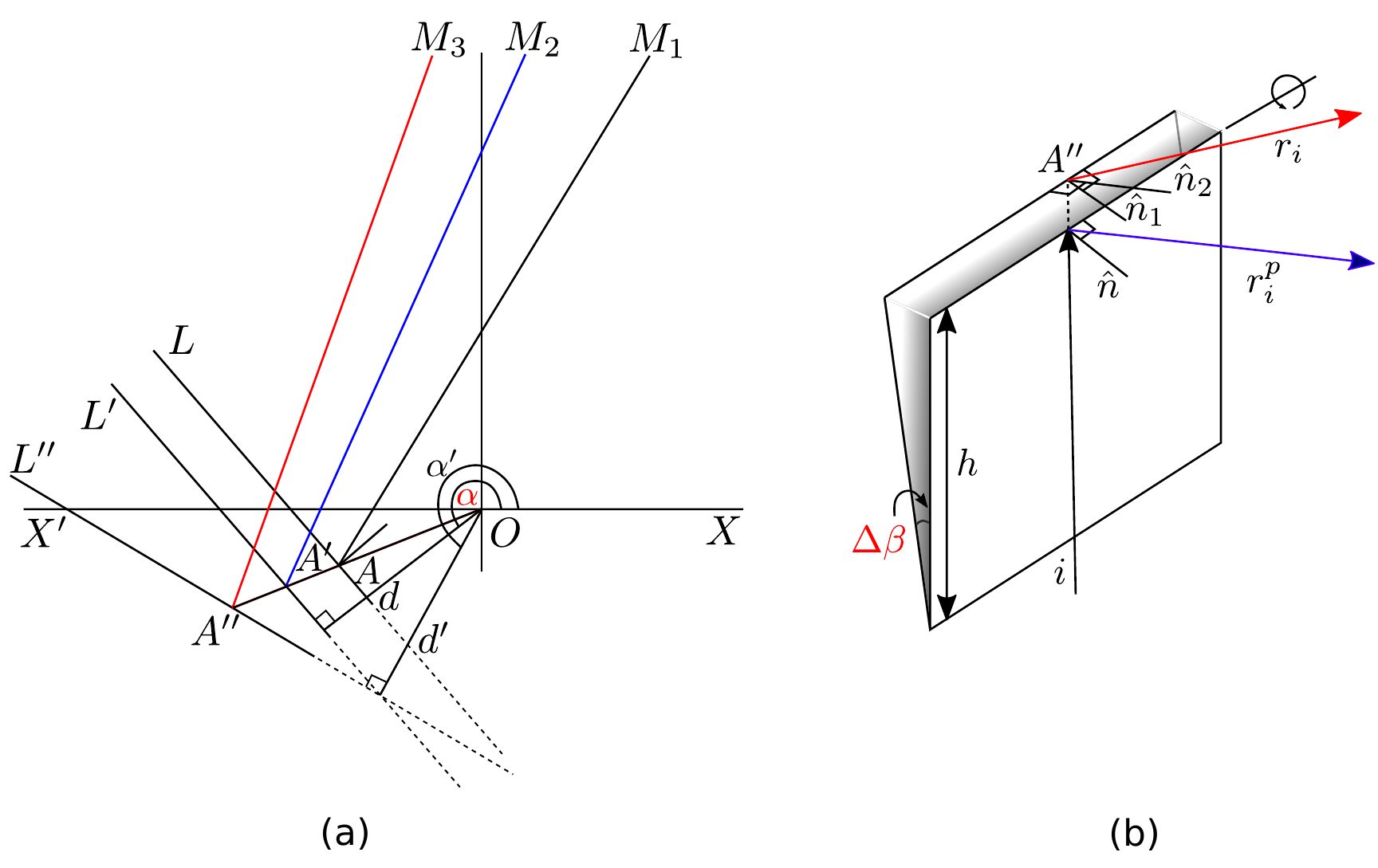}
		\end{center}
		\caption{Geometric description of perturbation in mirror placement; laser beams in (a) are in the scan plane affected by perturbation in $\alpha$ and $d$, (b) are further  perturbed by $\Delta\beta$ due to mirror tilt.}
		\label{fig:08}
	\end{figure}

	The construction of any design deviates from ideal parameter settings. The sensor designed in the present context may potentially be affected due to mechanical defects, which is reflected in the measured range information. As a result, the reconstructed depth image does not show the exact range and bearing as expected due to reflection of laser beams, as derived in the last section. Therefore, we need to calibrate the sensor by modeling small perturbations in the design parameters and obtain optimal perturbation in the design parameters to rectify the performance and achieve better shape reconstruction of an object. We identified three independent sources of perturbation, namely (a) horizontal orientation error $\Delta \alpha$ of the mirror, (b) perpendicular distance $\Delta d$ from the center of the sensor to the mirror, and c) mirror tilt $\Delta \beta$ about the axis of rotation of the mirror. Geometric definition of the perturbation parameters are demonstrated in Figure~\ref{fig:07}. Let’s consider, $r_i$ be the measured range due to reflecting mirror. The height of the mirror from the axis of rotation is denoted by $h$, and $\theta_i$ be the  bearing angle of $i$-th laser beam. The deviation of the mirror from $L$ to $L'$ causes the perpendicular distance to shift from $d$ to $d'$. In conjunction with this, any perturbation $\Delta \alpha$ in $\alpha$ would cause $L'$ to rotate and situate itself at $L''$ adding perturbation in orientation. This is demonstrated in Figure~\ref{fig:08}a. A reflected beam, which is expected to travel along $AM_1$, travels along $A''M_3$ due to the perturbation in $\alpha$ and $d$. The mirror could further tilt about its axis of rotation while settled after lift up, thereby brings about change in the direction of normal to the mirror from $\hat{n}$ to $\hat{n}_2$, shown in the Figure~\ref{fig:08}b. The beam $A''M_3$ will further be elevated by an amount of $2\Delta \beta$ due to the mirror tilt of $\Delta \beta$ and travel along $r_i$ as shown in Figure~\ref{fig:08}b.
	
	From the geometry of Figure~\ref{fig:07}a, $A''M_3$ corresponds to $r_i^p$ in Figure~\ref{fig:07}b, i.e. the component of the actual reflected ray $r_i$ projected on to the Lidar’s scan plane can be expressed as,
	\begin{equation}\label{eq:04}
	r_i^p = r_i -[(h \tan \Delta \beta + d + \Delta d) \tan(\alpha + \Delta \alpha - \theta_i)] \cos(2\Delta \beta)
	\end{equation}
	Note that $\Delta \beta$ can become either positive or negative and the height of the mirror is long enough to accommodate for the signal reflection due to small tilt  $\Delta \beta$. Now, based on the formulations carried out in Section III, we calculate the actual range and bearing of the line joining the measured location and the optical center of the sensor with the help of the measured range $r_i$. The actual projected range $r_i^a$ of the measured point from the sensor origin, by taking projection of the line joining the optical center $O$ and the measured location on the scan plane, and corresponding bearing angle $\theta_i^a$ with respect to the sensor's local coordinate system can be expressed as
	\begin{dmath}\label{eq:05}
		r_i^a = [(r_i^p - (h \tan \Delta \beta + d + \Delta d)\cos(2\Delta \beta))^2 + \\ (h \tan \Delta \beta + d + \Delta d) \tan(\alpha + \Delta \alpha - \theta_i)]^{1/2}
	\end{dmath}
	
	\begin{dmath}\label{eq:06}
		\theta_i^a = \bigg[\tan^{-1}\bigg\{\frac{r_p - (h \tan \Delta \beta + d + \Delta d)}{(h \tan \Delta \beta + d + \Delta d)\tan(\alpha + \Delta \alpha - \theta_i) }\bigg\} + \bigg(\alpha + \Delta \alpha - \theta_i - \frac{\pi}{2}\bigg) \bigg]
	\end{dmath}
	
	where, $\alpha$ and $d$ are known constants, $\beta = \frac{\pi}{2}$. Then, the coordinate of the position on an object, hit by the laser beam in the Lidar’s coordinate frame is given as,
	\begin{equation}\label{eq:07}
	x_i^l = r_i^a \sin \theta_a
	\end{equation}
	\begin{equation}\label{eq:07}
	y_i^l = r_i^a \cos \theta_a
	\end{equation}
	\begin{equation}\label{eq:08}
	z_i^l = r_i^p \sin(2\Delta \beta)
	\end{equation}
	The position vector in the local frame is then transformed to global fixed frame and expressed as,
	
	\begin{equation}\label{eq:09}
	{\mathbf{v_i}^l} = {\mathbf{R}^{-1}} {\mathbf{T}^{-1}} {\mathbf{v_i}^l}
	\end{equation}
	where, ${\mathbf{v_i}^f}$ = $\begin{bmatrix} x_i^l & y_i^l & z_i^l\end{bmatrix}^T$ is the vector in local frame and ${\mathbf{v}}_i^f$ is corresponding vector in the global frame. ${\mathbf{R^{−1}}}$ and ${\mathbf{T^{−1}}}$ are the inverse rotation and translation matrices respectively.
	
	As mentioned in section III, the placement of mirrors allows for one-third of the angular scan window to be exposed free from external mirror reflection. To estimate the optimal perturbation in the designed sensor, we require ground truth information, and the unreflected measurements aid to that. To make the calibration steps simple, we face the sensor in front of a reasonably planar and vertical wall and acquire range measurements. The measurements obtained from the unreflected region of the sensor are then passed through a principal component analysis (PCA) based plane fitting module, that provides the geometric description of the fitted plane. Now, for every measured location obtained from the un-calibrated sensor, there  exists a point on the fitted plane. Similarly, for every measurement due to any reflected laser beam, there should be a unique point on the fitted plane. This can be calculated from the derivation in section III. Let us represent $i$-th desired point vector as ${\mathbf{v}}_i^d$. Then, with the help of \eqref{eq:05} and \eqref{eq:06}, we formulate a cost function, by minimizing that we can obtain the optimal perturbation parameters.
	
	\subsection{Optimization steps}
	With reference to Figure~\ref{fig:07}, the proposed cost function is given by, 
	
	\begin{dmath}\label{eq:10}
		\argmin_{\Delta \alpha, \Delta \beta, \Delta d} \quad  \mathop{\sum_{k=1}^{N} \sum_{i=1}^{N_j}} |({\mathbf{v}_i^f} - {\mathbf{v}_i^d})\cdot\hat{{\mathbf {n}}}|^2               \\
		\text{s.t.} \quad \Delta \alpha \in (\alpha_1, \alpha_2), \Delta \beta \in(\beta_1, \beta_2), \Delta d \in (d_1, d_2)     \notag 
	\end{dmath}
	
	We used nonlinear least square method to optimize the cost function in \eqref{eq:10}. The sensor contains four mirrors and the perturbation parameters are independent for each mirror. So, we replicate same optimization process for all four mirrors. $N_j$ denotes the number of laser samples to be considered for $j^{th}$ mirror. For the optimization problem, we basically setup a nonlinear least square problem that can be solved using standard techniques such as gradient descent or steepest descent approach. It is to be verified that the optimized parameter settings are not biased to a specific orientation of the plane with respect to the sensor and that it provides same performance irrespective of any arbitrary orientation of the plane. Therefore, we introduce another variable $N$ in the outer summation that indicates the measured data acquired from the number of different planes. To check for degeneracy, $N=3$ has been necessarily chosen here. 
	
	\subsection{Calibration performance} 
\begin{figure*}[t]
		\centering
		\subfloat[]{
			\includegraphics[height = 1.8in, width=0.32 \linewidth]{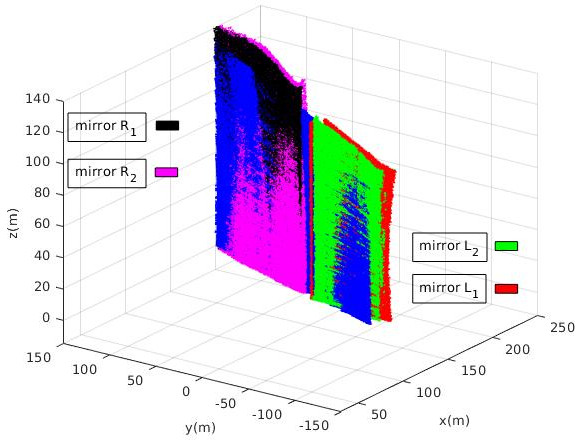}}
		\label{1a}\hfill
		\subfloat[]{
			\includegraphics[height = 1.85  in, width=0.32 \linewidth]{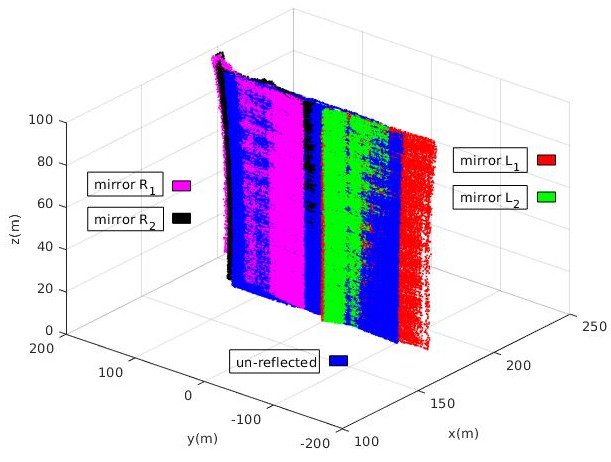}}
		\label{1b}\hfill
		\subfloat[]{%
			\includegraphics[height = 1.8in, width=0.321 \linewidth]{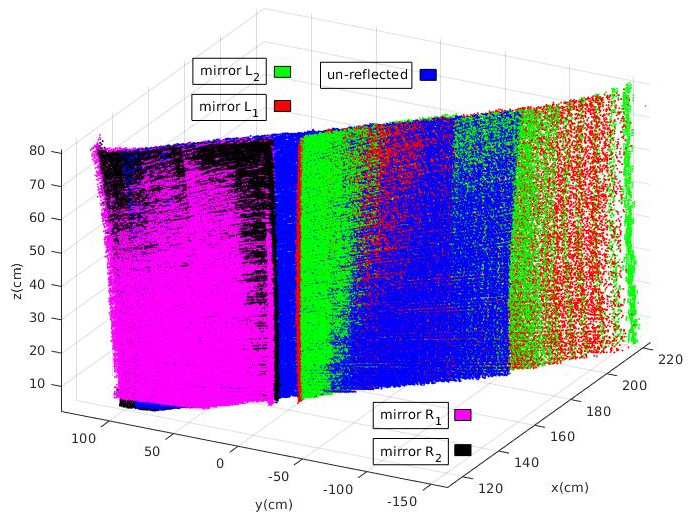}}
		\label{1a}\hfill \\
		\subfloat[]{
			\includegraphics[height = 1.6in, width=0.33 \linewidth]{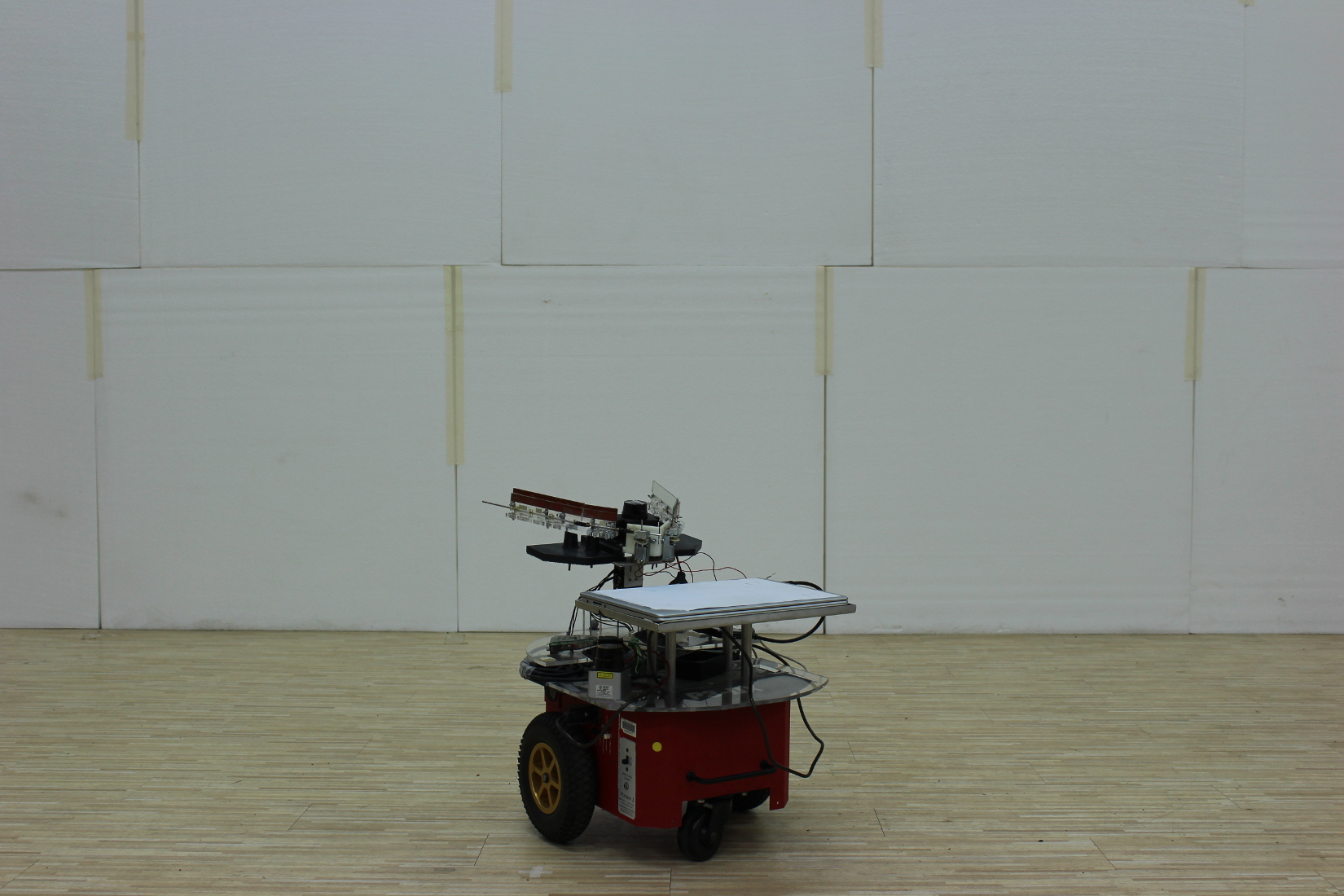}}
		\label{1a}\hfill
		\subfloat[]{
			\includegraphics[height = 1.6in, width=0.324 \linewidth]{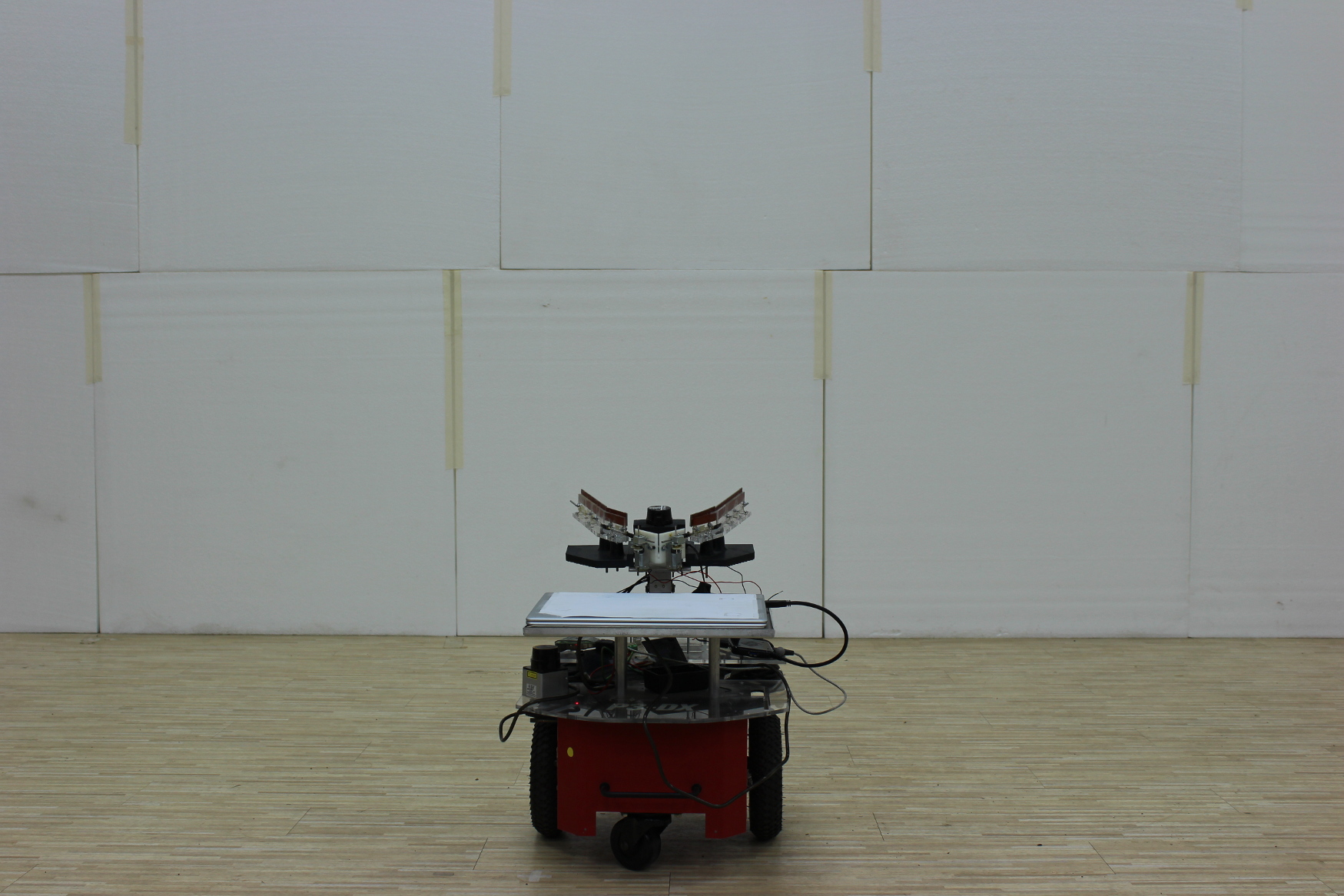}}
		\label{1b} 
		\subfloat[]{%
			\includegraphics[height = 1.6in, width=0.325 \linewidth]{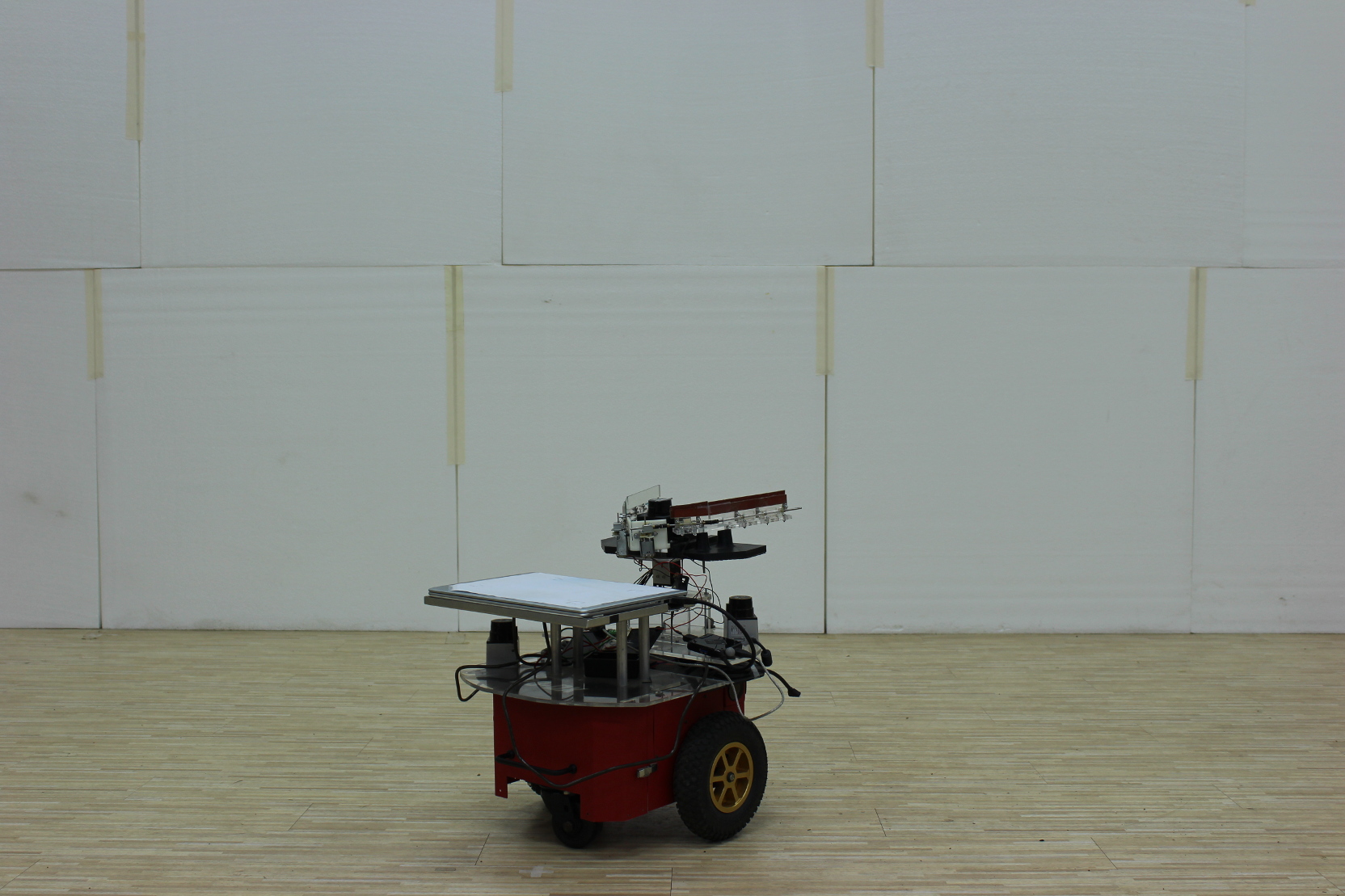}}
		\label{1a}
		\caption{Calibration performance is shown while placing the sensor facing a flat wall at three different orientations; (a) calibrated range measurement,
			while placing the sensor with orientation($\psi>\frac{\pi}{2}$) shown in (d); (b) calibrated performance while sensor is oriented ($\psi = \frac{\pi}{2}$)
			as shown in (e); (c) calibrated range perception for sensor orientation ($\psi < \frac{\pi}{2}$) as shown in (f).}
		\label{fig:09}
	\end{figure*}

	As a first step, we demonstrate the calibration performance of the sensor by optimally estimating the perturbation parameters incurred during the sensor development. From the results shown in Figure~\ref{fig:09}(a-c), the unreflected ground truth measurements are shown in blue whereas, reflected signals due to $R_1$ and $R_2$ are shown in violet and black respectively. Similarly, the signals due to $L_1$ and $L_2$ post calibration are shown in red and green respectively. We performed the experiment for three different orientations of the sensor facing the same scene, to obtain the optimal values of them. Typical Lidars show performance tolerance in range measurement. The specific Lidar chosen in this experiment shows $3\%$ range error. In spite of that, mirror reflected measurements after calibration is found to contain them within 4.5cm ($\pm 2.25cm$), which is promising.
	
	{\em Note:} Due to planar wall being the object here, while the sensor is oriented towards left($\psi < \frac{\pi}{2}$), the left side mirrors find wider visibility in the same angular window, resulting in wider spatial coverage exceeding the un-reflected one. The same happens for the right side mirrors while oriented twards right ($\psi > \frac{\pi}{2}$). Optimal perturbation parameters have been computed by solving the optimization problem in \eqref{eq:10} and are provided in Table~\ref{tab:01}.
	
	
	\begin{table}[t]
		\centering
		\caption{Perturbation Parameters}
		\begin{tabular}{c|c|c|c}
			\hline
			\hline
			mirror & $\alpha(rad) $ & $\Delta\beta (rad)$ & $d (m)$ \\
			\hline
			$L_1$ & -0.043 & 0.05 & 0.01 \\
			\hline
			$L_2$ & -0.03 & 0.035 & -0.012 \\
			\hline
			$R_1$ &  0.004 & -0.008 & 0.007 \\
			\hline
			$R_2$ & 0.03 & -0.01 & -0.005  \\
			\hline\hline
		\end{tabular}
		\label{tab:01}
	\end{table}
	
	\section{Dynamic Obstacle Handling}
	In this section, we perform a series of experiments based on nodding 2D Lidar, both without and with using mirror assembly. This explores the applicability of the sensor in dynamic environment, while fitted on a moving vehicle. Due to the operating principle of the sensor, it is understood that it traces a linear ramp while nodding in the actuation space \cite{ral18}. It means that the 3-D view of an object is generated over a number of scans, and for a dynamic object, its accurate motion estimation is coupled with high quality and reliable shape reconstruction. In view to that, we adapted the entropy minimization algorithm discussed in \cite{ijrr1}, a simple yet effective mechanism for moving object tracking by motion compensation. 
	
	\subsection{Pre-processing}
	As the sensor perceives an entire environment comprising of both static and dynamic objects, we need to segment out the measurements corresponding to the static objects. Due to sparse nature of measurements, we obtain the voxel representation of the entire environment.  Dynamic obstacles usually represent inconsistent measurements, unlike the static objects due to the relative motion between the sensor and obstacle(s). These inconsistencies in measurements help segmenting the moving objects from the static ones. As the name suggests, these measurements result in blurred shape  of the dynamic object in space and time. An optimal estimator is necessary to compute the state of the obstacle accurately, that will eventually produce a reliable shape of such an obstacle.

	
	\subsection{Motion Estimation}
	The motion estimation algorithm in this work enables a mobile robot to track a moving object. Here, the estimation algorithm is performed in two steps. Initially, we estimate the velocity and orientation coarsely so that we can develop a good initial guess that is close to the optimal state of the object. For that, we compute the centroid of each set of inconsistent scans (projected onto XY plane) corresponding to dynamic obstacle(s). The transformation between centroid pair in two consecutive scans provides a way to compute motion parameters in each scan interval.
	
	With a good initial estimate of the speed and orientation of the dynamic obstacle, we accumulate the measured cloud in each scan and represent them as an ensemble of fine grained voxels. For a poor final estimate of the motion parameters, a large number of voxels will generate resulting in blurred reconstruction of the obstacle. On the other hane, the optimal estimate of the motion parameters will result in the best shape reconstruction, as well as the number of voxels to fit in the motion compensated point cloud will be the least. The accumulated scans $S_a$ up to $n$ scans can be expressed as,
	\begin{equation}\label{eq:11}
	S_a = \bigcup_{k=1}^{n} \prod_{i=1}^{n-k} {\mathbf{T}_i} S_k
	\end{equation}
	where, ${\mathbf{T}_i}$ is the transformation between $(k-1)$-th and $k$-th inconsistent cloud, and $S_k$ denotes $k$-th such cloud. As discussed above, we need to define a metric that can be minimized to obtain the optimal motion parameters. Map entropy used as a quality assurance metric in \cite{ijrr1}, is found to be very effective in a way that the best map will correspond to the minimum entropy with respect to a blurred map with significantly higher value of it, for the same environment. As a result, the smallest number of voxels will be enough to reconstruct the best map. In this case, every time there is an inconsistent set of measurements, we compute the total entropy of the accumulated voxels $S_a$ starting from the first observation until that respective scan. Considering two possible states for any voxel, either occupied or unoccupied, the total entropy $H_t$ is given by,
	\begin{equation}\label{eq:12}
	H_t = -\sum_{i=1}^{N_v} \sum_{j=1}^{2} p_j \log p_j
	\end{equation}
	where, $p_j$ denotes the probability of occupancy of a voxel. By minimizing $H_t$ from the cost function in \eqref{eq:13}, we obtain the optimal transformations between each pair of inconsistent cloud and therefore, find out the optimal motion parameters, which also produces minimum voxel representation $N_v^*$ of the object. As time progresses and we acquire more and more measurements and the estimation accuracy improves  towards the convergence with the desired one. 
	
	\begin{equation}\label{eq:13}
	N_v^* = \argmin_{N_v} H_t
	\end{equation}
	
	\section{Experiments \& Results}
\begin{figure}[t]
		\centering
		\subfloat[]{
			\includegraphics[height = 1.5in, width=0.49 \linewidth]{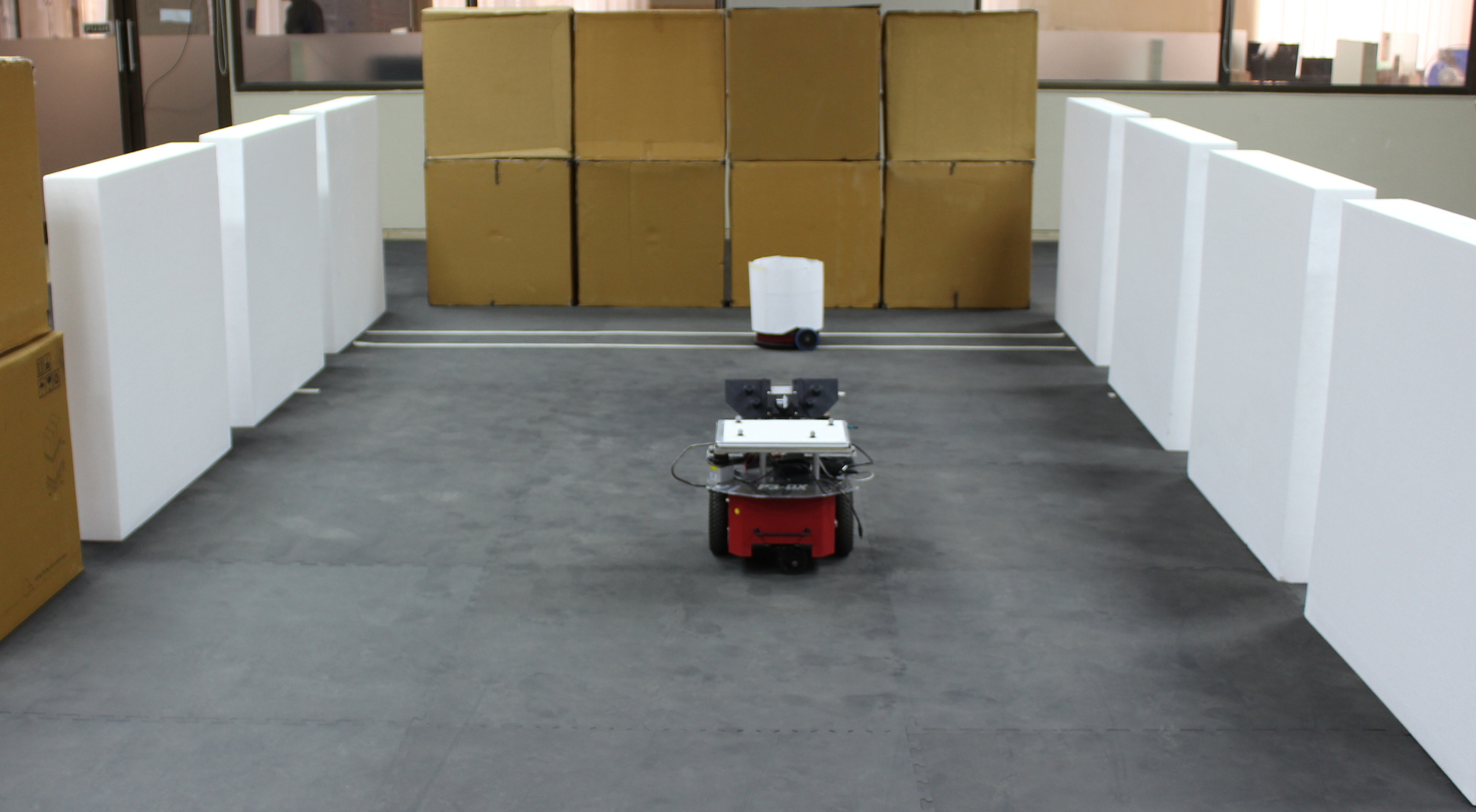}}
		\label{1a}\hfill
		\subfloat[]{
			\includegraphics[height = 1.5 in, width=0.48 \linewidth]{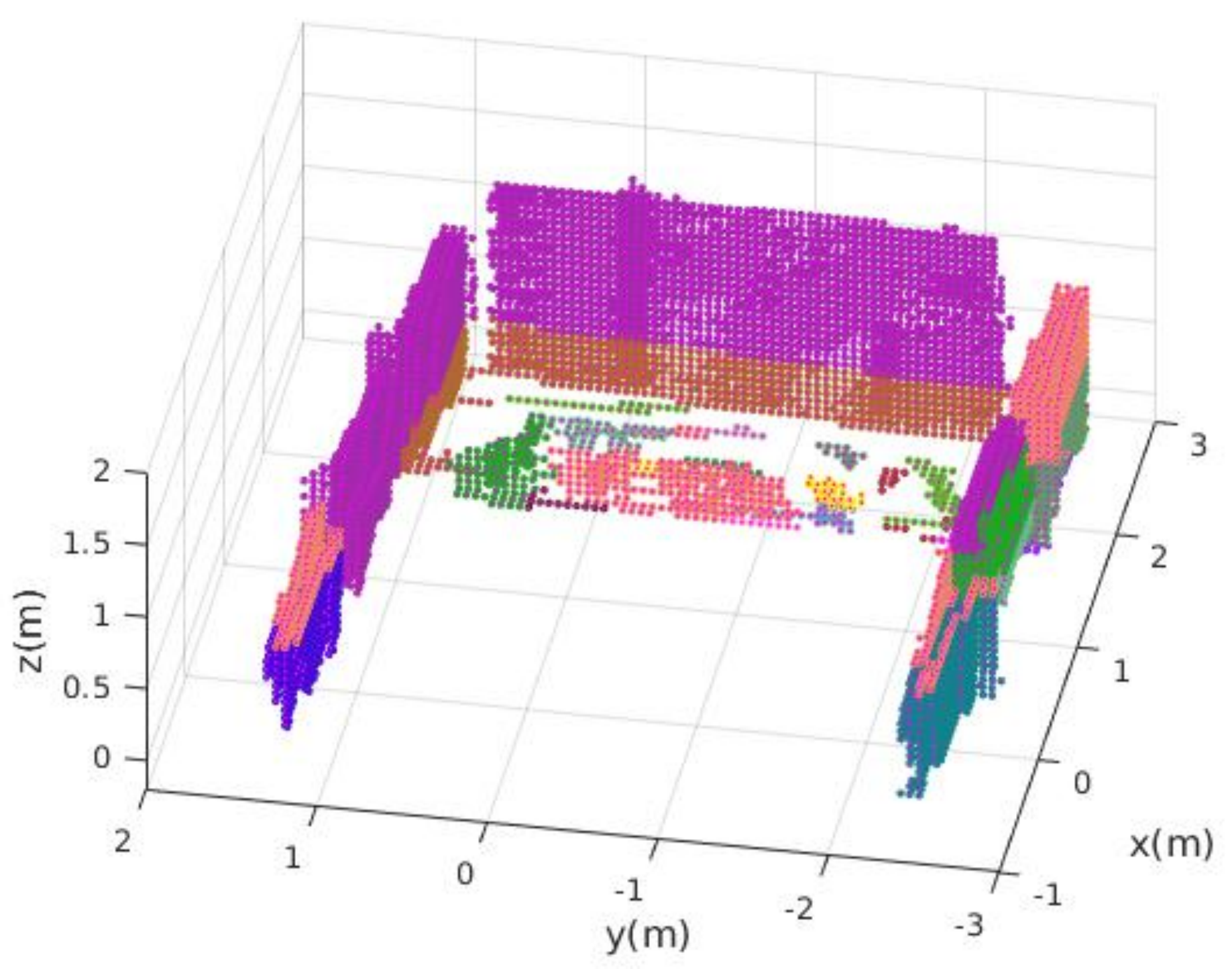}}
		\label{1b}\hfill \\
		\subfloat[]{
			\includegraphics[height = 1.2 in, width= \linewidth]{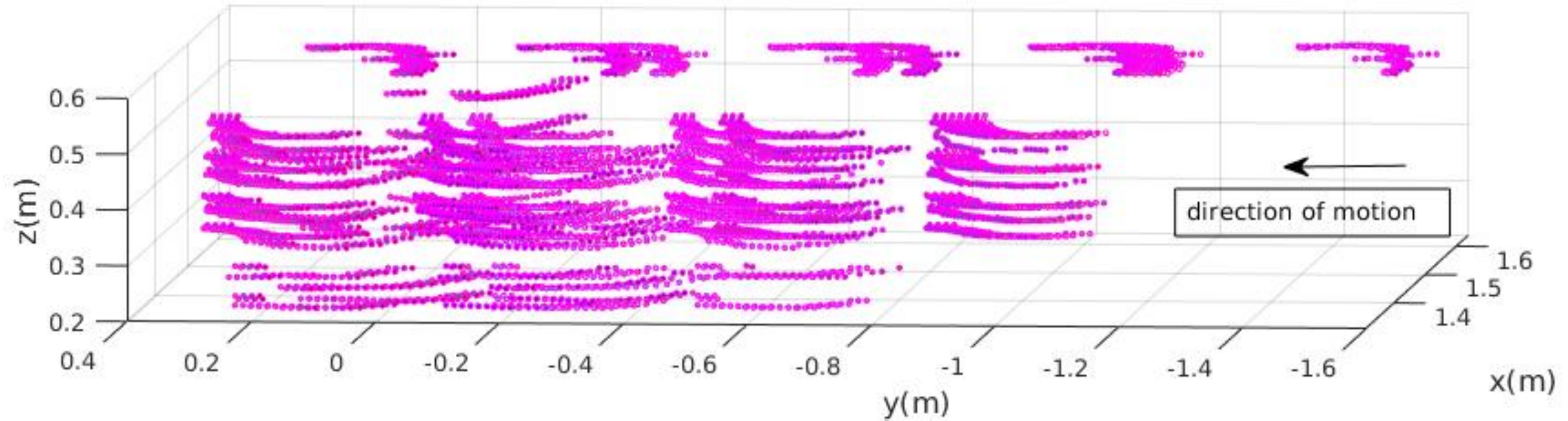}}
		\label{1b}\hfill\\
		\subfloat[]{
			\includegraphics[height = 1.2 in, width= \linewidth]{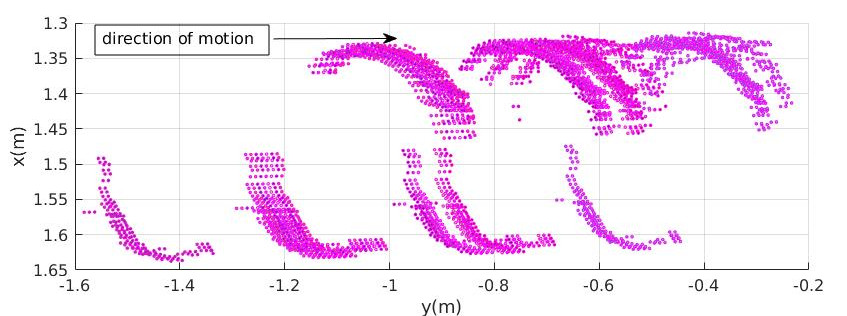}}
		\label{1b}
		\caption{A mobile robot in an indoor dynamic environment(a); Point-cloud segmentation from disjoint obstacles can be identified by different color of the segmented cloud, and inconsistent region is shown in the middle with more than one colors (b); time series representation of motion compensated accumulated obstacle cloud side view (b) and top view (c) are shown for without mirror case, with highly accurate reconstruction.}
		\label{fig:11}
	\end{figure}

	We set up experiments to test the motion estimation algorithm proposed in the previous section. For this purpose, we set up an indoor environment with a mobile robot Pioneer P3-DX carrying the proposed sensor and use another mobile robot acting as a dynamic object. Here, we assume that P3-DX carrying the sensor is facilitated with reliable localization. For simplicity, the moving object is wrapped with a white cylindrical structure to make it look like a moving cylinder as shown in Figure~\ref{fig:11}a. We allow the obstacle to run with a constant speed in a given direction and move back and forth.
	
\begin{figure}[t]
		\centering  
		\subfloat[]{
			\includegraphics[height = 1.4 in, width= \linewidth]{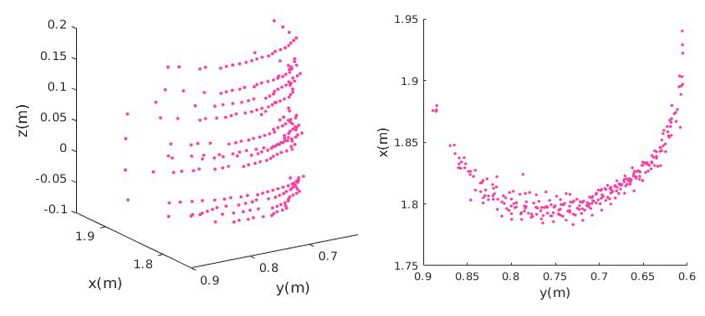}}
		\label{1b}\hfill \\
		\subfloat[]{
			\includegraphics[height = 1.4 in, width= \linewidth]{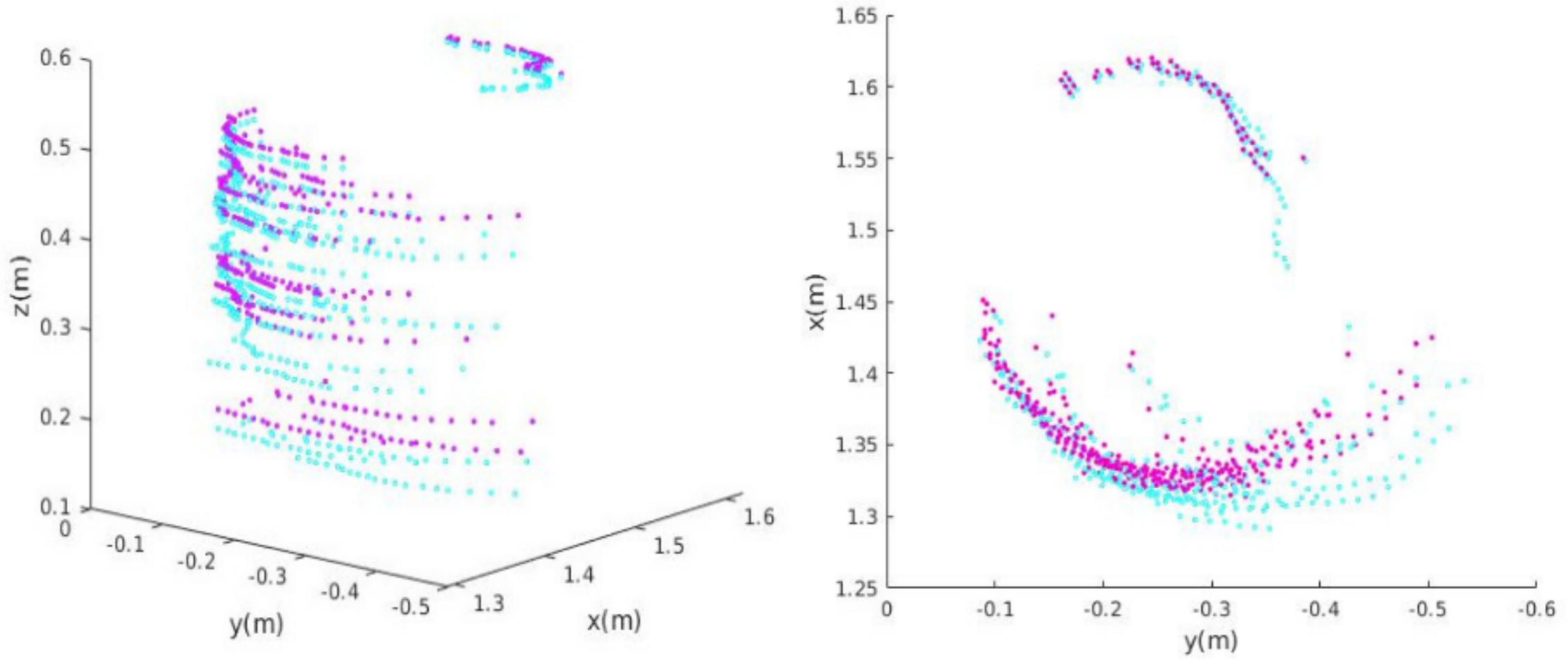}}
		\label{1b}\hfill
		\caption{(a) accumulated cloud after motion estimation (side and top view) without using mirror set-up; (b) when mirrors are lifted up and corresponding accumulated cloud with accurate motion estimation (side and top view). Each contiguous measurements are referred as a {\em trace} here.}
		\label{fig:10}
	\end{figure}

	Based on a good initial guess, once the motion parameters start to converge we show the accumulated cloud representation at a given time in Figure~\ref{fig:10}. In this result, we provide a qualitative as well as partially quantitative performance comparison between the two cases, without using mirrors and while using them. We acquire total 38 Lidar scans in each case out of which 13 scans have been able to trace the moving object in Figure~\ref{fig:10}a, whereas, for the same 13 scans we got to produce 28 traces of the obstacle while using the mirrors. The side view and the top view of the accumulated cloud are quite consistent in both the cases in qualitative sense. Although there is a little mis-alignment observed in the mirror reflected signals. In Figure~\ref{fig:11}, we show the detailed result. Segmented point cloud in (b) has been colored with different colors based on whether any two objects are found spatially disjoint by the sensor. Spatio-temporal occupancy of the dynamic obstacle is obtained with the help of a ray tracing algorithm that identifies any change in voxel occupancy probability. Using this information, we segment the traces of the dynamic obstacle with same color, when they are close enough based on the neighborhood classification metric used for segmentation. We show the time series representation of the accumulated cloud from both, the side and top view, after compensating the motion of the obstacle.
	
\begin{figure}[t]
		\centering
		\subfloat[]{
			\includegraphics[height = 1.5in, width=0.48 \linewidth]{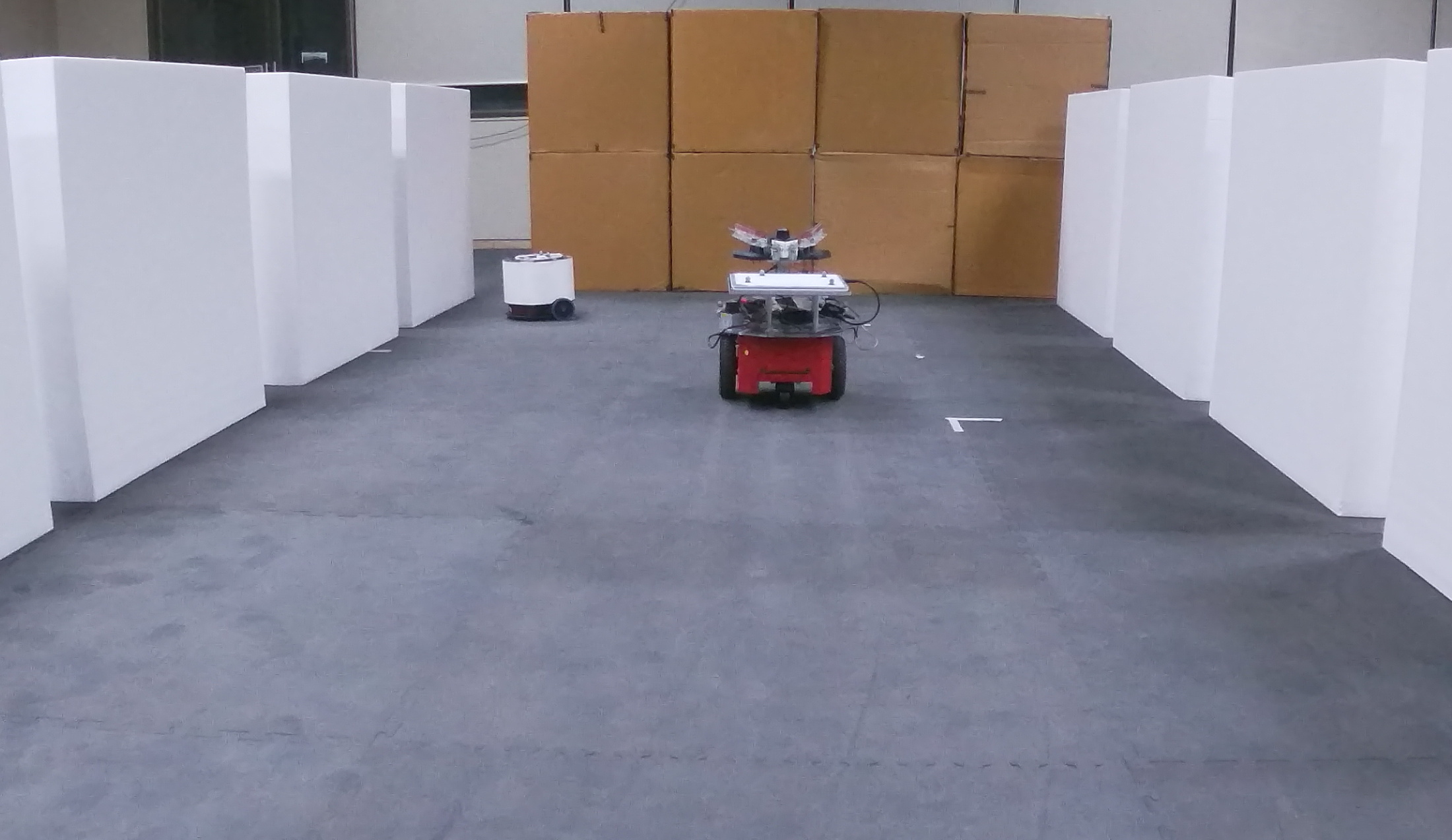}}
		\label{1a}\hfill
		\subfloat[]{
			\includegraphics[height = 1.7 in, width=0.49 \linewidth]{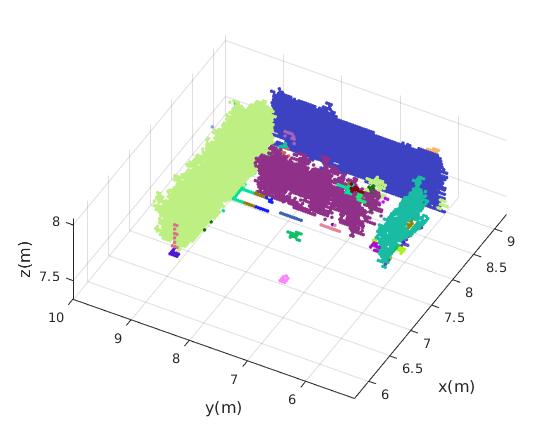}}
		\label{1b}\hfill \\
		\subfloat[]{
			\includegraphics[height = 1.3 in, width= \linewidth]{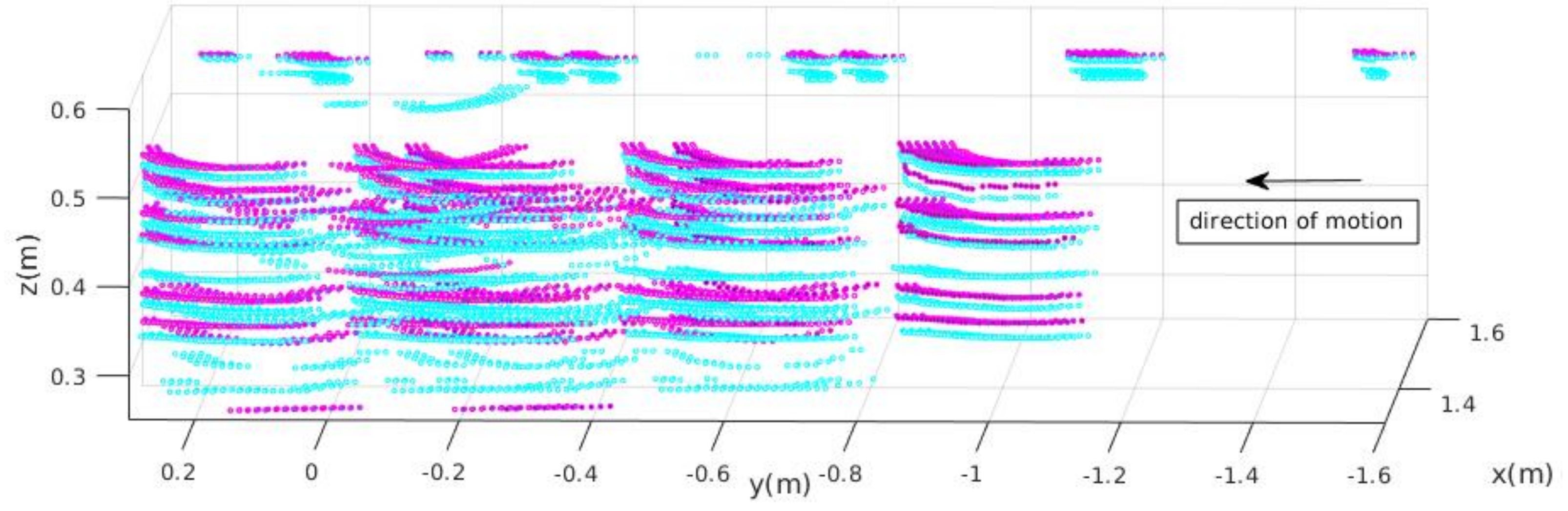}}
		\label{1b}\hfill\\
		\subfloat[]{
			\includegraphics[height = 1.3 in, width= \linewidth]{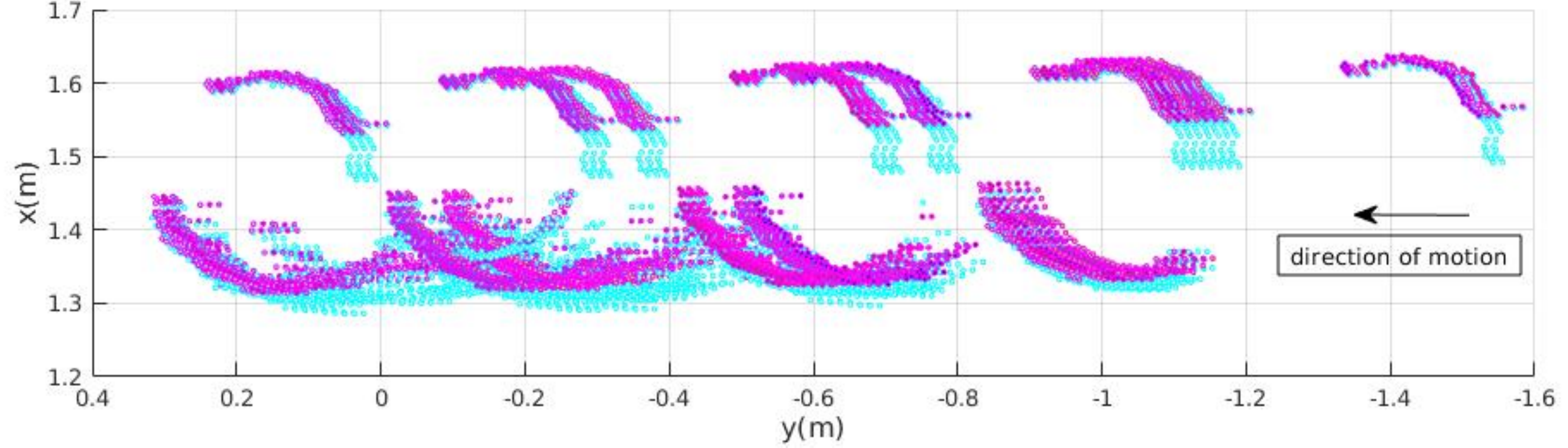}}
		\label{1b}
		\caption{(a) Similar experimental set up as used in Figure~\ref{fig:12}(a), except the proposed sensor used here; (b) disjoint static obstacles can be identified by different color of the segmented cloud, a lot more inconsistent region appears in the middle compared to as the previous case; side view (c) and top view (d) of the reconstructed obstacle after motion estimation; magenta shows measurements from un-reflected region and cyan represents the same due to reflecting mirrors.}
		\label{fig:12}
	\end{figure}

\begin{figure}[t]
		\begin{center}
			\includegraphics[height=2.9cm, width=7.8cm]{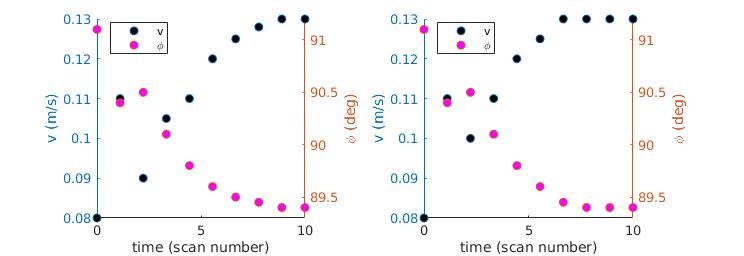}
			\caption{convergence of motion parameters velocity (v) and orientation ($\phi$); (right) converges faster while using mirror set-up, compared to the (left) without using mirrors.}
			\label{fig:13}
		\end{center}
	\end{figure}

	While Figure~\ref{fig:11} shows the result  without using the extrinsic mirrors, the results in Figure~\ref{fig:12}, reflect the performance of the proposed sensor while using these mirrors. While using the mirrors, we not only find that the measurements are concentrated within a narrow angular window, the proposed sensor also acquires a lot more spatially dense measurements at the same time. This is evident from Figure~\ref{fig:11}b  while comparing with Figure~\ref{fig:12}b from qualitative point of view. Total 600 scans have been used for  each setup. While viewing the motion compensated  time series measurements we find that for the same number of Lidar scan input, the proposed sensor with the help of mirrors brings out more spatial information of the obstacle compared to in case without mirrors. As motion estimation is the matter of interest, we provide a comparison of convergence of the motion estimation performance in Figure~\ref{fig:13}. We find that the mirror set-up does help in achieving convergence of the estimated motion parameters faster than while using its natural setting. As for example, the ground truth motion parameters of the obstacle in either way, captured by the indoor positioning system VICON is $(v, \phi) = (0.13m/s, 89.4^{\circ})$. Result show that it takes 9 scans to achieve convergence when the mirrors are not used, whereas the same is observed to happen from scan number 7 on-wards, while using mirrors. 
	
	\subsubsection*{Future Work}	
	There is further scope to improve the design of this sensor. The mirror base on each side of the Lidar is currently rigidly fixed, which can be made flexible by providing  motorized control along the yaw. It will help generate different type of laser beam interference pattern and allow to choose required yaw angle of the mirror bases based on desired scan density. From another view point, one could look for different actuation mechanisms that may not require DC motors resulting in reduced noise and weight of the sensor. Miniaturization of the mirrors remains as a live problem for this sensor. We assumed that the mirrors are optically flat. It can be relaxed and considered in the sensor calibration procedure to help improving the sensor measurements.
	\section{Conclusion}
	In this paper, we propose a novel reconfigurable 3-D Lidar. It can be useful for both wide FoV navigation as well as tracking dynamic object(s) within a narrow angular region with three fold denser measurements. A novel calibration mechanism has been proposed for the sensor which also shows typically dense measurement that can be expected from this sensor. To prove its efficacy in real world robotics problems, it was mounted on a mobile robot and has been used to detect and track a dynamic obstacle. It shows promising results by acquiring more than two fold measurements compared with no-mirror case and the shape reconstruction performance also goes in line in comparison with the earlier.    
	
	\bibliographystyle{IEEEtran}
	\bibliography{thesis.bib}
\end{document}